\documentclass{article}
\usepackage{ijcai22}

\usepackage{times}
\usepackage{soul}
\usepackage{bm}
\usepackage{url}
\usepackage[hidelinks]{hyperref}
\usepackage[utf8]{inputenc}
\usepackage[small]{caption}
\usepackage{graphicx}
\usepackage{amsmath}
\usepackage{amsthm}
\usepackage{booktabs}
\usepackage{algorithm}
\usepackage{algorithmic}
\usepackage{amsfonts}
\usepackage[table,xcdraw]{xcolor}
\usepackage{pifont}
\usepackage{adjustbox}
\usepackage{multirow}
\usepackage{booktabs}
\usepackage{graphicx}
\usepackage{subfigure}
\usepackage[table]{ xcolor}
\usepackage{hyperref}
\hypersetup{
    colorlinks=false,
	bookmarks=true,
	bookmarksopen=false,
	bookmarksnumbered=true,
    linkcolor=[rgb]{0 0 1},
    filecolor=[rgb]{0 0 1},      
    urlcolor=[rgb]{0 0 1},
    citecolor=[rgb]{0 0 1},
}
\makeatother

\definecolor{lightGreen}{RGB}{203, 231, 202}
\title{Hyperparameter-free and Explainable Whole Graph Embedding}

\author{
Hao Wang$^{1,2}$\footnote{These authors contributed equally to this work.}\and
Yue Deng$^{1,2}$\footnotemark[1]\and
Linyuan Lü$^{1,2,3}$\footnote{Corresponding Author: linyuan.lv@uestc.edu.cn}\And
Guanrong Chen$^4$\\
\affiliations
$^1$University of Electronic Science and Technology of China\\
$^2$Yangtze Delta Region Institute (Huzhou), University of Electronic Science and Technology of China\\
$^3$Beijing Computational Science Research Center\\
$^4$City University of Hong Kong\\
\emails
}

\begin{document}

\maketitle

\begin{abstract}
Graphs can be used to describe complex systems. Recently, whole graph embedding (graph representation learning) can compress a graph into a compact lower-dimension vector while preserving intrinsic properties, earning much attention. However, most graph embedding methods have problems such as tedious parameter tuning or poor explanation. This paper presents a simple and hyperparameter-free whole graph embedding method based on the DHC (Degree, H-index, and Coreness) theorem and Shannon Entropy (E), abbreviated as DHC-E. The DHC-E can provide a trade-off between simplicity and quality for supervised classification learning tasks involving molecular, social, and brain networks. Moreover, it performs well in lower-dimensional graph visualization. Overall, the DHC-E is simple, hyperparameter-free, and explainable for whole graph embedding with promising potential for exploring graph classification and lower-dimensional graph visualization.
\end{abstract}

\section{Introduction}
\label{Introduction}

Graph (networks) can be used to describe the connections or associations between objects, which are ubiquitous in our daily lives. Examples include social networks, citation networks, knowledge graphs, brain networks, etc. Graph analysis helps explore hidden information in complex systems and implement link prediction, node classification, clustering \cite{vasconcelos2021segregation}, visualization \cite{Janssens2022DecodingGR}, and for understanding brain disorders. In graph analysis, it is observed that some large graphs are with small sparsity, where each node has only a few edges, resulting in a long nodal feature vector with many zeros \cite{Barabsi2009ScaleFreeNA}, thereby causing traditional graph analytic methods high computation cost. To address this issue, some graph embedding methods have been proposed \cite{Perozzi2014DeepWalkOL,grover2016node2vec}, which learn and compress the node or whole graph features to a lower-dimensional vector and maximally preserve the intrinsic graph properties.

Before 2000, the main effort on network features learning was dimension reduction of higher-dimensional data. Traditional methods include principal component analysis (PCA), linear discriminant analysis (LDS), multiple dimensional scaling (MDS), etc. However, they are linear methods, which do not always perform well when there are nonlinear relationships within datasets. Subsequently, some other dimension reduction methods came about around 2000, among which popular ones were isometric mapping (IsoMap) \cite{Tenenbaum2002TheIA} and Laplacian eigenmaps (LLE) \cite{Belkin2003LaplacianEF}. Both IsoMap and LLE were manifold learning. From a more practical perspective, in 2013 word2vec \cite{Mikolov2013DistributedRO} was developed using the SkipGram model to learn lower-dimensional vector representations of words (word embedding). The excellent performance of word2vec promoted a wave of activities on embedding learning in scientific communities. Based on the output features, one may classify the current graph embedding methods into four categories: node embedding \cite{cao2015grarep}, edge embedding \cite{bordes2013translating}, substructure embedding \cite{yanardag2015deep}, and whole graph embedding \cite{narayanan2017graph2vec}. 

The current study focuses on \textit{whole graph embedding}. Several whole graph embedding methods have been proposed. A typical one is the family of graph spectral distances (FGSD), which calculates the Moore-Penrose spectrum of the normalized Laplacian and uses the histogram of the spectral features as a whole graph representation \cite{Verma2017HuntFT}. Additionally, Graph2vec \cite{narayanan2017graph2vec} algorithm uses unsupervised methods to derive fixed-length task-agnostic embeddings of graphs and shows good performance in graph classification. Invariant graph embedding (IGE) computes a mixture of spectral and node embedding based features and pools node feature embedding to create graph descriptors \cite{galland2019invariant}. Graph and Line graph to vector (GL2vec) complements either the edge label information or the structural information which Graph2vec missed \cite{chen2019gl2vec}. 

However, to date, the aforementioned advanced approaches are suffering some issues: 
\begin{itemize}
\item Mathematically intractable, sophisticated, and challenging to inspire beginners.
\end{itemize}
\begin{itemize}
\item Costly hyperparameter tuning process oriented to different data sets, without definite interpretability.
\end{itemize}
\begin{itemize}
\item Moreover, these algorithms are lacking evaluation on biological systems like brain networks.
\end{itemize}
Thus, it is desirable to provide simple, hyperparameter-free, easy-to-understand methods to make the graph embedding applicable for interdisciplinary researches. Here, we present a network compression and representation method by integrating a generalized DHC theorem and the Shannon entropy (E), abbreviated as DHC-E. Specifically, the network's intrinsic properties can be compressed into a one-dimensional entropy sequence, and each order of entropy represents the scale information carried by the corresponding DHC iteration at that order. As the DHC iterations gradually and automatically converge to coreness, the amount of information they correspond to is gradually reduced. Our \textbf{main contributions} are summarized as follows:
\begin{enumerate}
\item  We present a simple and hyperparameter-free graph embedding method, abbreviated as DHC-E, bracing its generalization into more interdisciplinary researches.
\item  We provide Python, MATLAB, and Julia codes of the DHC-E with detailed instructions in GitHub. 
\item  Overall, the DHC-E shows comprehensive advantages compared with three state-of-the-art whole graph embedding methods for network classification, stability and time consuming. 
\item  Furthermore, We apply the DHC-E to brain networks in distinguishing autism participants from typical controls, extending the experiment to brain systems and make the brain morphological network available as a candidate benchmark dataset for future researches.
\end{enumerate}

\begin{table}[h]
\scriptsize
\begin{adjustbox}{center} 
\begin{tabular}{|lcccc|}
\toprule
\textbf{Key Features}  & {\textbf{DHC-E}} & GL2vec & IGE & Graph2vec         \\
\midrule
Simple model & {\color[HTML]{F56B00} \ding{51}}& 
{\color[HTML]{3166FF} \ding{55}}  & 
{\color[HTML]{3166FF} \ding{55}}  & 
{\color[HTML]{3166FF} \ding{55}} \\
Hyperparameter-free & {\color[HTML]{F56B00} \ding{51}} & 
{\color[HTML]{3166FF} \ding{55}} & 
{\color[HTML]{3166FF} \ding{55}} & 
{\color[HTML]{3166FF} \ding{55}} \\
Brain morphological network & {\color[HTML]{F56B00} \ding{51}}& 
{\color[HTML]{3166FF} \ding{55}}& 
{\color[HTML]{3166FF} \ding{55}}& 
{\color[HTML]{3166FF} \ding{55}} \\
Classification performance & {\color[HTML]{F56B00} 1st} & 2nd & 3rd   & 4th   \\
Stability  & 2nd & 3rd  & {\color[HTML]{F56B00} 1st} & 4th \\
Time complexity & 2nd & 4th & 3rd   &{\color[HTML]{F56B00} 1st}   \\
\bottomrule
\end{tabular}
\end{adjustbox} 
\caption{Overall comparison between different models. {\color[HTML]{F56B00} \ding{51}} indicates Yes; {\color[HTML]{3166FF} \ding{55}} indicates No.}
\label{Overall_comparison}
\end{table}

We examine its accuracy and effectiveness by comparing it with three state-of-the-art whole graph embedding methods, using modeled network, small molecules network, social network, brain network, and computer vision network, etc. Our DHC-E method shows a better trade-off between complexity and good performance than baseline methods on various benchmark datasets.
% The remaining of the article is organized as follows: In section \ref{Sec2}, we present some preliminaries, including a short introduction to graph, DHC metric, Shannon Entropy, and whole graph embedding. In section \ref{Sec3}, we present the derivation of DHC-E and apply different datasets to test the effectiveness of our method on graph classification tasks. We utilize three state-of-the-art effective complex models that at least need two hyperparameters as our baseline. In section \ref{Sec4}, we use both simulated and empirical data to test the graph classification performance of our model against three baseline models. We also evaluate the stability and time complexity of all models. Furthermore, we combine our model with each baseline to test whether integration features can improve the graph classification performance. In addition, we use the DHC-E features to demonstrate an exploratory application in lower-dimension graph visualization. In section \ref{Sec5}, we provide a summary and present a vision for developing simple and effective graph embedding methods.

\section{Preliminaries}
\label{Sec2}
In this section, some preliminaries are presented, preparing for the technical development of the paper.
\subsection{Graph}
A graph is denoted by $G=(V,E)$, where $V$ is the set of nodes and $E \subseteq (N \times N)$ is the set of edges. The edges of $G$ can be weighted or unweighted, and directed or undirected. This paper only considers \textit{unweighted and undirected graphs} with no self-loops or multiple edges.

\subsection{Degree, H-index and coreness}
\label{degree-H-index-coreness}
A node's \textit{degree} can measure its influence in the graph: the higher the degree is, the more nodes are connected with, so the more influence it has. A node's \textit{H-index} (short for \textit{Hirsch index}) \cite{hirsch2005index} is the maximum value $h$ such that it has at least $h$ neighbors with degree no less than $h$ \cite{lu2016h}. A node's \textit{coreness} further takes its location in the graph into account, measuring its influence based on the $k$-core decomposition process \cite{dorogovtsev2006k}, where a larger coreness indicates that a node locates more centrally in the graph. See \textbf{\textit{Supporting Information (SI) Sec. A1}} for details for the calculation of the three metrics.

\subsection{Shannon entropy} 
Without loss of generality, for a random variable $X$, suppose it has $n$ possible values with probabilities $p_i$ $(i=1,2,...,n)$ correspondingly. The Shannon entropy \cite{shannon1948mathematical} of $X$ is measured by

\begin{equation}
H=-\sum_{i=1}^n p_i \log_2 p_i. 
\label{Shannon_entropy_formula}
\end{equation}

\subsection{Whole graph embedding}
The \textit{whole graph embedding} \cite{zhang2018network} of a graph is a vector derived from a mapping $\Phi: \mathcal{G} \to \mathbb{R}^{k}$, which compresses a graph $\mathcal{G}$ to the vector space $\mathbb{R}^{k}$ ($k$ is the vector dimension). For the classification of different graphs, such vectors are supposed to preserve graph properties as much as possible to distinguish graph categories.

\section{Methods}
\label{Sec3}

Sec.~\ref{subsec_DHC-E} introduces DHC-entropy (DHC-E), and its performance on graph classification tasks are described in Sec.~\ref{performance_evaluation}. 
\subsection{DHC-entropy (DHC-E)}
\label{subsec_DHC-E}

Since the DHC theorem reveals nodal H-index sequences in the graph, we develop a whole graph embedding method, abbreviated as \textit{DHC-entropy (DHC-E)}, by combining these sequences as nodal features and the Shannon entropy, which is simple, hyperparameter-free, and explainable.
% In this section, the DHC theorem is reviewed in Sec.~\ref{DHC_theorem} and the Shannon entropy of the DHC-E is introduced in Sec.~\ref{DHC_entropy}. Finally, the performance of DHC-E is compared with three state-of-the-art baselines in Sec.~\ref{performance_evaluation} using different evaluation metrics and datasets. 

\subsubsection{DHC theorem}
\label{DHC_theorem}
L{\"u} et al. \cite{lu2016h} revealed the relationship among degree, H-index and coreness, described as the \textit{DHC theorem}: a node's H-index is calculated based on the degree centrality of its neighbors by the process illustrated in Sec.~\ref{degree-H-index-coreness}. Following the same process, its H-index can be updated based on the previous H-indices of its neighbors. Such updating process continues iteratively, producing an H-index sequence of the node. However, the sequence will not be infinite because the DHC theorem proves its convergence to the node's coreness.

More specifically, denote degree of node $i$ as $k_i$, and its neighbors' degree as $k_{j_1}, k_{j_2}, ..., k_{j_{k_i}}$. Let $\mathcal{H}$ denote the operation of calculating (updating) H-index of node $i$. Then, it becomes
\begin{equation}
h_i=\mathcal{H}(k_{j_1}, k_{j_2}, ..., k_{j_{k_i}}).    
\end{equation}
Note that $k_{j_1}, k_{j_2}, ..., k_{j_{k_i}}$ are sorted by descending order. Generally, it denotes node $i$'s zero-order H-index and first-order H-index as $k_i$ and $h_i$, respectively. Then, the calculation of its $n$th-order H-index ($n>0$) can be described as 
\begin{equation}
\begin{aligned}
&h_i^{(n)}=\mathcal{H}\left(h_{j_1}^{(n-1)}, h_{j_2}^{(n-1)}, ..., h_{j_{k_i}}^{(n-1)}\right),\, 
\\ &\operatorname{where}\, h_i^{(0)}=k_i, h_i^{(1)}=h_i.
\end{aligned}
\end{equation}
The DHC theorem proves that node $i$'s H-index sequence $h_i^{(0)},h_i^{(1)},h_i^{(2)},...$ will converge to its coreness $c_i$, i.e.,
\begin{equation}
c_i = \lim_{n \to \infty} h_i^{(n)}.
\end{equation}
The same is true for other nodes in the graph. See \textbf{\textit{SI. Sec. A2}} for details.

\subsubsection{Combining Shannon entropy and DHC theorem}
\label{DHC_entropy}

Inspired by the idea of Weisfeiler-Lehman kernel \cite{shervashidze2011weisfeiler} representing a graph by integrating nodal labels in each iteration of a convergent updating process, our proposed DHC-E aims to employ the nodal H-index sequences illustrated in Sec.~\ref{DHC_theorem} to learn the graph embedding. To complete this, two essential issues must be addressed: (1) How to integrate the nodal H-index sequences; (2) The constructed whole graph embedding of different graphs is usually different in dimensions. How to automatically align them.

For the first point, suppose the DHC updating process on a graph $G=(V,E)$ with $n$ nodes (i.e., $|V|=n$) converges after $s$ iterations, producing $n$ H-index sequences $\{(h_i^{(0)},h_i^{(1)},...,h_i^{(s)})|i=1,2,...,n \}$, where $h_1^{(m)},h_2^{(m)},...,h_n^{(m)}$ $(0 \leq m \leq s)$ implies $G$'s state in the $m$-th iteration. To represent the state $m$, DHC-E first calculates the probability distribution of $h_1^{(m)},h_2^{(m)},...,h_n^{(m)}$, and then puts it into Formula (\ref{Shannon_entropy_formula}) in order to measure its Shannon entropy $H_m$, which quantifies the uncertainty of the graph at state $m$. See \textit{\textbf{SI. Sec. A3-4}} for details.

For the second issue, as the iterations of the above DHC updating process on different graphs are usually unequal in dimensions (see \textit{\textbf{SI. Sec. A4}}), the resulting whole graph embeddings are consequently incompatible for some downstream machine learning tasks (like graph classification). To align them, DHC-E further takes the largest dimension of all produced embeddings as the baseline and any other with lower dimension will be replenished with its last element.

In this transductive process, the rationale behind DHC-E is better explainable compared with most hyperparameter tuning-based machine learning methods. 

\subsection{Performance evaluation}
\label{performance_evaluation}

We evaluated DHC-E's performance using three state-of-the-art methods as baselines, and performed plenty of graph classification tasks by K-nearest neighbor (KNN) algorithm.

\subsubsection{Datasets}
\label{datasets}
Several experimental datasets, including \textit{simulation datasets} and \textit{real-world datasets} (See \textit{\textbf{SI. Sec. B}} for details), are employed to comprehensively evaluate the performance of DHC-E in different scenarios as following.

\begin{figure}[]
\centering
\includegraphics[scale=0.9]{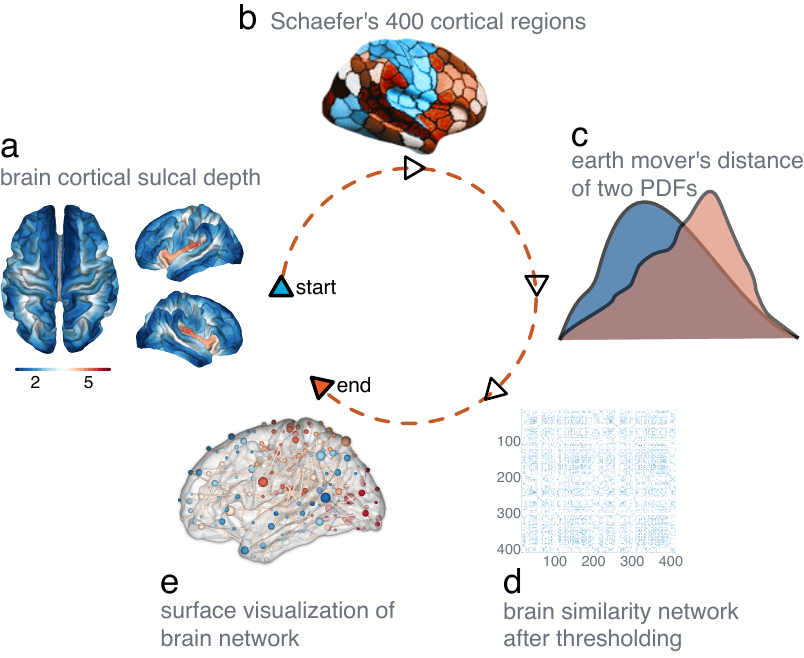}
\caption{\textbf{Flowchart of constructing the brain sulcal depth similarity network.} \textbf{a}, First obtain the sulcal depth for each node in the brain surface space. \textbf{b}, Then, use the Schaefer 400 atlas to define the brain nodes. \textbf{c}, For each brain region, estimate their PDF of sulcal depth and calculate the earth mover's distance between any paired brain regions, resulting in a $400 \times 400$ similarity matrix with top 10\% edges preserved (\textbf{d}), Convert the distance metric to similarity metric through sigmoid function. \textbf{e}, Surface plot of the brain sulcal depth similarity network.}
\label{brain flowchart}
\end{figure}

\textit{Datasets for binary classification}. SW and BA were used to construct \textit{categorized datasets by simulation} with distinctive complexity. For real-world datasets, TUDatasets \cite{Morris+2020} about multiple domains are used to combine \textit{sub-categorized datasets from the real world}. Brain morphological similarity networks \cite{Wu2018} were generated for 20 autism spectrum disorders (ASD) and  20 typical control (TC) participants (Figure. \ref{brain flowchart}). The datasets were labelled B1-B8, see \textit{\textbf{SI. Sec. B.1}} for details.

\textit{Datasets for multi-class classification.} 
To explore the ability of DHC-E in multi-class classification tasks, following the above strategies, datasets M1-M8 (See \textit{\textbf{SI. Sec. B.2}} for more details) were constructed.

\subsubsection{Baseline Methods}
\label{baselines}
We compared DHC-E with the three baselines:
\begin{itemize}
\item GL2vec \cite{chen2019gl2vec}: GL2vec: Graph Embedding Enriched by Line Graphs with Edge Features. 
\item IGE \cite{galland2019invariant}: Invariant Embedding for Graph Classification.
\item Graph2vec \cite{narayanan2017graph2vec}: Graph2vec: Learning Distributed Representations of Graphs.
\end{itemize}

\subsubsection{Evaluation metrics}
Accuracy (ACC) and F1 scores (F1) \cite{riesen2010graph} are used to quantify the performance of graph classification results. The better the performance of graph classification is, the larger the values of ACC and F1 will be.

\subsubsection{Graph classification}
Given a graph set $\mathbb{G}=\{G_1, G_2, ..., G_n\}$, where $G_i=(V_i,E_i,l_i)$ with node set $V_i$, edge set $E_i$ and label $l_i$ denoting its category, implementing DHC-E on each $G_i$ in $\mathbb{G}$ produces $n$ whole graph embeddings as
\begin{equation}
\{(H_{i,1}, ...,H_{i,s_i})|i = 1,2,..,n\},
\end{equation}
where $H_{i,j}$ is the Shannon entropy of $G_i$ in the $j$-th iteration within the total $s_i$ convergence steps of the DHC updating process. Then, these embeddings are replenished with their last elements to align in dimension $s_{\operatorname{max}}$. Finally, an $n \times s_{\operatorname{max}}$ matrix is constructed as the input of KNN classifier. After comparing with other classifiers such as support vector machine (SVM), KNN performed more efficiently and still effectively. The pseudo code of DHC-E for graph classification is shown in Algorithm 1.
\begin{table}[h]
\scriptsize
\begin{adjustbox}{center}
\begin{tabular}{l}
\toprule
\textbf{Algorithm 1.} DHC-E for graph classification \\ \hline
\textbf{Input:} A set of $n$ graphs $\mathbb{G}=\{G_1,G_2,...,G_n\}, G_i=(V_i,E_i,l_i)$.\\
\textbf{Output:} the classification results $l_1',l_2',...,l_n'$ of $\mathbb{G}$.\\

\textbf{1\,\,:} DHC\_EntropyMatrix = [\,]: \\
\textbf{2\,\,:} \textbf{for} each $G_i=(V_i,E_i,l_i)$ in $\mathbb{G}$: \\
\textbf{3\,\,:} \quad Hi = [$d_1,d_2,...,d_{\operatorname{|V_i|}}$]\\
\textbf{4\,\,:} \quad $H_{i,0}$ = ShannonEntopy(Hi)\\
\textbf{5\,\,:} \quad j=1:\\
\textbf{6\,\,:} \quad \textbf{while} true:\\
\textbf{7\,\,:} \quad \quad HiUpdated = DHC\_UpdatingProcess (Hi)\\
\textbf{8\,\,:} \quad \quad \textbf{if} HiUpdated is equal to Hi:\\
\textbf{9\,\,:} \quad \quad \quad break\\
\textbf{10:} \quad \quad \textbf{else:}\\
\textbf{11:} \quad \quad \quad $H_{i,j}$ = ShannonEntropy(HiUpdated)\\
\textbf{12:} \quad \quad \quad Hi = HiUpdated \\
\textbf{13:} \quad \quad \quad $j = j+1$\\
\textbf{14:} \quad add $[H_{i,0}, H_{i,1},...]$ to DHC\_EntropyMatrix by row\\
\textbf{15:} DHC\_EntropyMatrix = DimensionAlignment(DHC\_EntropyMatrix)\\
\textbf{16:} $l_1',l_2',...,l_n'=$KNN(DHC\_EntropyMatrix)\\
\bottomrule
\end{tabular}
\end{adjustbox}
\end{table}

To ensure the fairness of performance comparison, the embeddings obtained from different models were taken as the input of a common KNN classifier with fixed hyperparameters for graph classification (see line 16 in Algorithm 1). For every model, its overall performance is averaged by 500 runs of the KNN classifier with 10-fold cross-validation.

\section{Experimental results}
\label{Sec4}
\begin{figure}[h]
\centering
\includegraphics[scale=0.6]{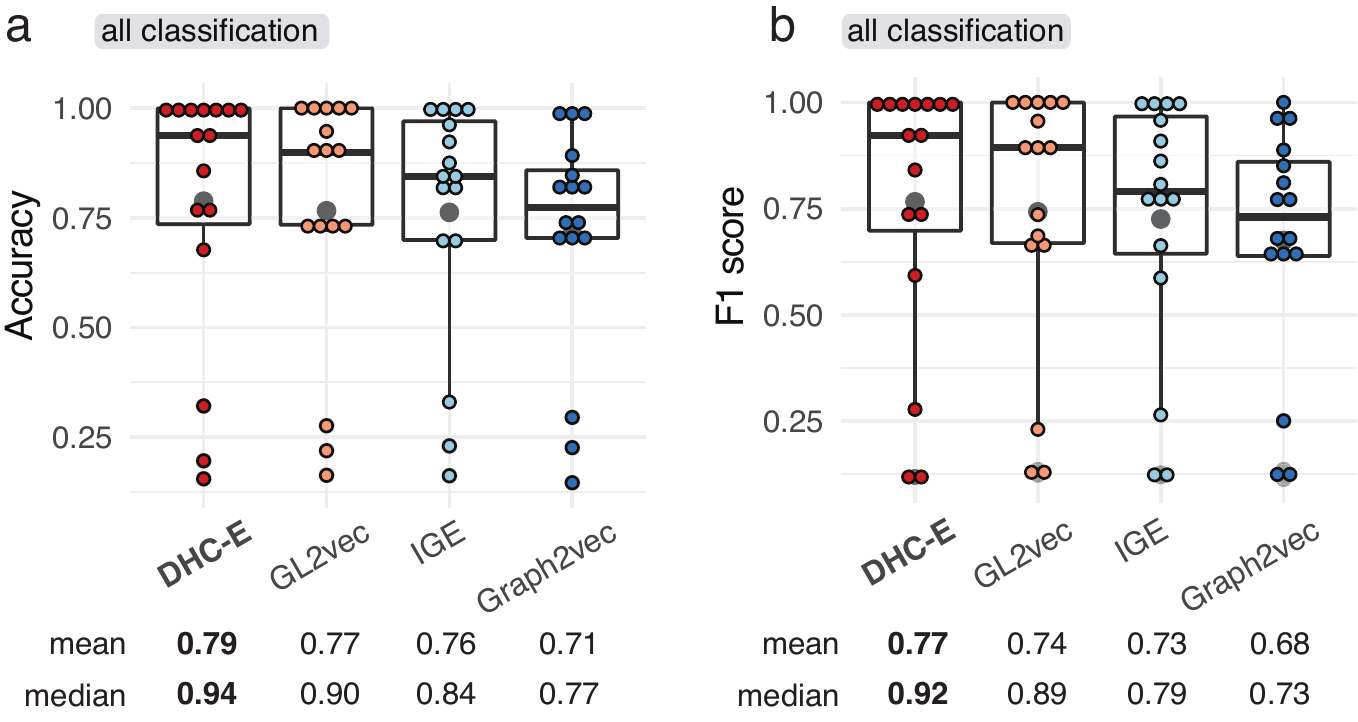}
\hspace{0.1in}
\caption{\textbf{Classification performance of different graph embedding models.} The plots were sorted by their mean values (in descending order). For all cases (B1-B8 and M1-M8), the accuracy and F1 score is presented in \textbf{a} and \textbf{b}, respectively. The black dots indicate the mean classification value of each model.}
\label{ClassifyResults} 
\end{figure}

The ACC and F1 results for DHC-E and baselines on binary and multi-class classification tasks are summarized in Figure.~\ref{ClassifyResults}, Table.~\ref{binary_classification_results} and Table.~\ref{multi-class_classification_results}, respectively, where the corresponding stability results are also shown. The time complexity results are presented in Table.~\ref{Time_complexity}. 

Overall, DHC-E achieves the first rank across all the datasets based on the mean classification value and provides a better trade-off between the accuracy, stability, and time complexity. Additionally, an application of DHC-E on graph visualization is shown in Figure.~\ref{visGraph}.

\subsection{Results on binary graph classification}
For binary graph classification, the ACC and F1 results for DHC-E and baselines are summarized in Table.~\ref{binary_classification_results}, in which the eight datasets are classified into three different subtasks for binary graph classification (i.e., the subtasks based on \textit{categorized datasets by simulation}, \textit{sub-categorized datasets from the real world}, and \textit{categorized datasets}), where the ACC and F1 results for evaluated models are averaged, respectively. The hyperparameter settings of baselines follow the default values in the python package, named \textit{karate club}, as shown in \textbf{\textit{SI. Sec. B.3}}. Each result is calculated by averaging over 500 runs on independently random split train-test sets. Clearly, DHC-E outperforms other models for all three subtasks of binary classification.

Specifically, in comparing the overall performance of DHC-E with the weakest baseline on the three subtasks, DHC-E outperforms Graph2vec on the first subtask (i.e., B1, B2, and B3) by 16.59\% and 22.69\% over ACC and F1, respectively. Regarding the second subtask (B4, B5, and B6), the ACC and F1 results of DHC-E are 6.64\% and 7.62\% higher than those of Graph2vec, respectively. Similarly, on the third task (i.e., B7 and B8) DHC-E outperforms GL2vec by 5.12\% and 5.37\% over ACC and F1, respectively. Clearly, the DHC-E has been demonstrated that even a simple model can achieve good classification performance, and in some cases even outperform complex models.

These results indicate that the difficulty of graph sets to be classified may be distinctive under specific scenarios. Thus, one should make a trade-off between accuracy and stability (see Tables.~\ref{binary_classification_results} and \ref{multi-class_classification_results}), and time consuming (see Table.~\ref{Time_complexity}), in order to select appropriate models for graph classification.

\begin{table*}[h]
\scriptsize
\begin{adjustbox}{center} 
\begin{tabular}{@{}ccccccccccc@{}}
\toprule
\multirow{2}{*}{\textbf{datasets}} & \multicolumn{2}{c}{\textbf{DHC-E}} & \multicolumn{2}{c}{\textbf{Graph2vec}} & \multicolumn{2}{c}{\textbf{IGE}} & \multicolumn{2}{c}{\textbf{GL2vec}} \\ \cmidrule(l){2-9} 
     & \textbf{ACC} & \textbf{F1} & \textbf{ACC}& \textbf{F1}& \textbf{ACC}& \textbf{F1}& \textbf{ACC}  & \textbf{F1}    \\ \midrule
B1   & 1.000 (0.000) & 1.000 (0.000) & 1.000 (0.000)  & 1.000 (0.000)  & 0.850 (0.000) & 0.807 (0.000) & 0.740 (0.032) & 0.674 (0.036) \\
B2   & 1.000 (0.000) & 1.000 (0.000) & 0.740 (0.028)  & 0.634 (0.033)  & 1.000 (0.000) & 1.000 (0.000) & 1.000 (0.000) & 1.000 (0.000)\\
B3   & 1.000 (0.000) & 1.000 (0.000) & 0.833 (0.029)  & 0.811 (0.036)  & 0.962 (0.000) & 0.958 (0.000) & 1.000 (0.000) & 1.000 (0.000)\\
\midrule
\textbf{Avg.}   & 1.000 (0.000) & 1.000 (0.000) & 0.858 (0.019)  & 0.815 (0.023) & 0.937 (0.000) & 0.922 (0.000) & 0.913 (0.011) & 0.891 (0.012)              \\
\textbf{\textit{Imp.}} & \textbf{-}  & \textbf{-} & \textit{\textbf{+16.59\%}} & \textit{\textbf{+22.69\%}} & \textit{\textbf{+6.67\%}} & \textit{\textbf{+8.49\%}} & \textit{\textbf{+9.48\%}} & \textit{\textbf{+12.20\%}} \\
\midrule
B4   & 0.857 (0.011) & 0.841 (0.012) & 0.704 (0.012)  & 0.692 (0.012) & 0.875 (0.003) & 0.862 (0.004) & 0.745 (0.018) & 0.736 (0.018) \\
B5   & 0.755 (0.014) & 0.733 (0.015) & 0.704 (0.031)  & 0.668 (0.039) & 0.701 (0.000) & 0.663 (0.000) & 0.717 (0.030) & 0.686 (0.037) \\
B6   & 0.677 (0.031) & 0.593 (0.036) & 0.738 (0.054)  & 0.653 (0.073) & 0.694 (0.057) & 0.587 (0.079) & 0.740 (0.000) & 0.654 (0.000) \\
\midrule
\textbf{Avg.}   & 0.763 (0.019) & 0.722 (0.021) & 0.715 (0.032)  & 0.671 (0.041) & 0.757 (0.020) & 0.704 (0.028) & 0.734 (0.016) & 0.692 (0.018)            \\
\textbf{\textit{Imp.}} & \textbf{-} & \textbf{-} & \textit{\textbf{+6.64\%}} & \textit{\textbf{+7.62\%}} & \textit{\textbf{+0.80\%}} & \textit{\textbf{+2.61\%}} & \textit{\textbf{+3.93\%}} & \textit{\textbf{+4.36\%}} \\
\midrule
B7   & 1.000 (0.000) & 1.000 (0.000) & 0.977 (0.005)  & 0.976 (0.005)  & 1.000 (0.000) & 1.000 (0.000) & 0.903 (0.011) & 0.898 (0.012) \\
B8   & 1.000 (0.000) & 1.000 (0.000) & 0.975 (0.006)  & 0.949 (0.014)  & 1.000 (0.000) & 1.000 (0.000) & 1.000 (0.000) & 1.000 (0.000) \\
\midrule
\textbf{Avg.}   & 1.000 (0.000) & 1.000 (0.000) & 0.976 (0.006)  & 0.962 (0.010) & 1.000 (0.000) & 1.000 (0.000) & 0.951 (0.006)  & 0.949 (0.006)              \\ 
\textbf{\textit{Imp.}} & \textbf{-} & \textbf{-} & \textit{\textbf{+2.46\%}} & \textit{\textbf{+3.92\%}} & \textit{\textbf{0.00\%}} & \textit{\textbf{0.00\%}} & \textit{\textbf{+5.12\%}} & \textit{\textbf{+5.37\%}} \\\bottomrule
\end{tabular}
\end{adjustbox} 
\caption{Results on binary graph classification tasks.}
\label{binary_classification_results}
\end{table*}

\subsection{Results on multi-class graph classification}

For multi-class graph classification tasks, the ACC and F1 results for DHC-E and baselines are summarized in Table.~\ref{multi-class_classification_results}, in which the eight datasets are classified into three different subtasks for multi-class graph classification (i.e., the subtasks based on \textit{categorized datasets by simulation}, \textit{sub-categorized datasets from the real world}, and \textit{categorized datasets}), where the ACC and F1 results for evaluated methods are averaged, respectively. The hyperparameter settings of baselines follow the default values in the python package, named \textit{karate club}, as shown in \textbf{\textit{SI. Sec. B.3}}. Each result is calculated by averaging over 500 runs with 10-fold cross validation. It shows that the overall advantages of DHC-E are still satisfactory on the three subtasks compared with that of baseline, but not as prominent as that on binary graph classification tasks.

Specifically, on the first subtask (i.e., M1, M2, and M3), although DHC-E is 1.32 \% and 1.84\% lower than GL2vec over ACC and F1, respectively, it still significantly outperforms IGE and Graph2vec, where the ACC and F1 results of DHC-E are 12.29\% and 15.94\% higher than those of IGE, and 23.34\% and 30.28\% higher than those of Graph2vec, respectively. As for the second subtask (i.e., M4, M5, and M6), DHC-E achieves the accuracy in the middle level among all models. On the third subtask (i.e., M7 and M8), DHC-E is slightly inferior in the evaluated models. Its ACC and F1 results are 3.83\% and 6.19 \% lower than that of the best-performed baseline (i.e., GL2vec), respectively.

Besides, the overall accuracy of all models on the second subtask (i.e., M4, M5, and M6) is much worse compared with that on the other two subtasks, as shown in Table.~\ref{multi-class_classification_results}. This result reveals that the effective method for multi-class classification remains to be a great challenge today.

\begin{table*}[h]
\scriptsize
\begin{adjustbox}{center} 
\begin{tabular}{@{}ccccccccccccc@{}}
\toprule
\multirow{2}{*}{\textbf{datasets}} & \multicolumn{2}{c}{\textbf{DHC-E}} & \multicolumn{2}{c}{\textbf{Graph2vec}} & \multicolumn{2}{c}{\textbf{IGE}} & \multicolumn{2}{c}{\textbf{GL2vec}} \\ \cmidrule(l){2-9} 
     & \textbf{ACC} & \textbf{F1} & \textbf{ACC}& \textbf{F1}& \textbf{ACC}& \textbf{F1}& \textbf{ACC}  & \textbf{F1}\\ \midrule
M1   & 1.000 (0.000) & 1.000 (0.000) & 0.820 (0.000)  & 0.775 (0.000) & 0.827 (0.000) & 0.771 (0.000) & 1.000 (0.000) & 1.000 (0.000)           \\
M2   & 0.950 (0.012) & 0.932 (0.018) & 0.704 (0.059)  & 0.640 (0.067) & 0.923 (0.000) & 0.909 (0.000) & 1.000 (0.000) & 1.000 (0.000)          \\
M3   & 0.925 (0.010) & 0.913 (0.012) & 0.807 (0.026)  & 0.768 (0.031) & 0.810 (0.000) & 0.774 (0.000) & 0.913 (0.001) & 0.898 (0.001)           \\
\midrule
\textbf{Avg.}   & 0.958 (0.007) & 0.948 (0.010) & 0.777 (0.028) & 0.728 (0.033) & 0.853 (0.000) & 0.818 (0.000) & 0.971 (0.000) & 0.966 (0.000)           \\
\textbf{\textit{Imp.}} & \textbf{-} & \textbf{-} & \textit{\textbf{+23.34\%}} & \textit{\textbf{+30.28\%}} & \textit{\textbf{+12.29\%}}  & \textit{\textbf{+15.94\%}} & \textit{\textbf{-1.32\%}} & \textit{\textbf{-1.84\%}} \\
\midrule
M4   & 0.321 (0.013) & 0.278 (0.013) & 0.295 (0.030)  & 0.251 (0.028) & 0.330 (0.000) & 0.265 (0.000) & 0.276 (0.029) & 0.231 (0.026)          \\
M5   & 0.155 (0.007) & 0.119 (0.006) & 0.146 (0.015)  & 0.114 (0.013) & 0.162 (0.000) & 0.120 (0.000) & 0.163 (0.014) & 0.124 (0.012)          \\
M6   & 0.196 (0.011) & 0.119 (0.008) & 0.226 (0.025)  & 0.136 (0.019) & 0.230 (0.000) & 0.128 (0.000) & 0.219 (0.024) & 0.136 (0.019)          \\
\midrule
\textbf{Avg.}   & 0.224 (0.010) & 0.172 (0.009) & 0.222 (0.024) & 0.167 (0.020) & 0.241 (0.000) & 0.171 (0.000)  & 0.219 (0.022)  & 0.164 (0.019)               \\
\textbf{\textit{Imp.}} & \textbf{-} & \textbf{-} & \textit{\textbf{+0.71\%}} & \textit{\textbf{+2.84\%}} & \textit{\textbf{-7.16\%}} & \textit{\textbf{+0.56\%}} & \textit{\textbf{+2.03\%}} & \textit{\textbf{+4.83\%}} \\
\midrule
M7   & 0.991 (0.002)  & 0.991 (0.002)  & 0.892 (0.010)   & 0.888 (0.011)   & 0.994 (0.000)  & 0.994 (0.000)  & 0.894 (0.011)  & 0.889 (0.012)           \\
M8   & 0.780 (0.005)  & 0.740 (0.003)  & 0.847 (0.009)   & 0.851 (0.010)   & 0.839 (0.000)  & 0.773 (0.000)  & 0.947 (0.006)  & 0.956 (0.005)           \\
\midrule
\textbf{Avg.}   & 0.885 (0.003)  & 0.866 (0.002)  & 0.870 (0.010)  & 0.870 (0.011)  & 0.916 (0.000)  & 0.883 (0.000)  & 0.921 (0.009) & 0.923 (0.009)            \\ 
\textbf{\textit{Imp.}} & \textbf{-} & \textbf{-} & \textit{\textbf{+1.84\%}} & \textit{\textbf{-0.47\%}} & \textit{\textbf{-3.38\%}} & \textit{\textbf{-2.01\%}} & \textit{\textbf{-3.83\%}} & \textit{\textbf{-6.19\%}}\\
\bottomrule
\end{tabular}
\end{adjustbox} 
\caption{Results on multi-class graph classification tasks.}
\label{multi-class_classification_results}
\end{table*}

\subsection{Stability analysis}
Stability, as opposed to the deviation from the averaged performance of a model, is also an important metric for assessment, for it quantifies the potential of a model to reach its best performance compared to baselines, which determines the cost of experiments.

The corresponding standard deviations of the ACC and F1 results for DHC-E and baselines on datasets for both binary classification and multi-class classification are summarized in Tables.~\ref{binary_classification_results} and \ref{multi-class_classification_results}, which reveals the stability of different methods over 500 runs with split train-test sets. For a model, the lower the averaged standard deviations of ACC and F1 results, the higher the model's stability. Tables.~\ref{binary_classification_results} and \ref{multi-class_classification_results} show the overall averaged standard deviations of ACC results on IGE, DHC-E, GL2vec, and Graph2vec are 0.004, 0.007, 0.011, 0.021, respectively, and the standard deviations of F1-score results on the four models are 0.005, 0.008, 0.011, 0.025, respectively, indicating that DHC-E has satisfactory stability among the evaluated models.

\subsection{Time complexity analysis}
Here, we systemically investigate the time complexity of DHC-E and three baseline models to make a more comprehensive understanding of the trade-off between accuracy and time complexity.

The corresponding consuming time of DHC-E and baselines on different datasets is summarized in Table.~\ref{Time_complexity}, in which the time consumption, measured in seconds, is averaged from 500 separate runs. DHC-E achieves the second-best efficiency in overall time complexity among all of the four models, as shown in Table.~\ref{Time_complexity}, resulting in 69.23\% higher time consumption than Graph2vec. However, DHC-E has 150.13\% and 1520.47\% lower time consumption than IGE and GL2vec, respectively.

The researcher can choose different models depending on the specific situation. It should be emphasized that our Python code of DHC-E is not explicitly optimized, and there is still some room for performance improvement. In addition, we also provide Julia and MATLAB code of DHC-E, both of which are fast and can complete the B6 task in \textbf{1.8s} and \textbf{4s}, separately, on a personal computer [MacBook Pro (Retina, 13-inch, Early 2015); Processor 2.7 GHz Dual-Core Intel Core i5]. 

%For more details, see the link (\url{https://github.com/HW-HaoWang/DHC-E}).
 
%We believe that the value of a method shouldn't be merely evaluated by its accuracy but also by its scalability. For instance, a method that could achieve optimal performance on large-scale datasets but at a high running cost has a low scalability because it is almost ineffective in large-scale datasets. In light of this point, we further evaluate the time consumption of DHC-E, TGM, DHC-E+TGM, and baselines on every dataset in order to explore their computational complexity. 

%Based on the results shown in Table \ref{Time_consumption}, when the data size is small, the difference in time consumption of each method is slight, and all of them are within an efficient range. When we increase the data size, graph2vec has an overall lower time consumption compared to other methods because it adopts a numerical optimization method for calculation. Specifically, on the datasets with a middle-scale, including B4-B6 and M4-M6, the time consumption of DHC-E is lower than TGM and sometimes even lower than graph2vec, which indicates that DHC-E can run fast in most cases. This is because DHC-E adopts iterative calculation, which is similar to numerical optimization in machine learning methodology and maintains explainability. However, we don't recommend applying DHC-E on large-scale datasets preferentially when there is a lack of sufficient computation resources, even though it can achieve better performance shown in Tables \ref{simulation_datasets_results} and \ref{real-world_datasets_results}.

\begin{table}[h]
\scriptsize
\begin{adjustbox}{center} 
\begin{tabular}{@{}cccccc@{}}
\toprule
\textbf{datasets} & \textbf{\begin{tabular}[c]{@{}c@{}}Time \\ (DHC-E)\end{tabular}} & \textbf{\begin{tabular}[c]{@{}c@{}}Time\\ (Graph2vec)\end{tabular}} & \textbf{\begin{tabular}[c]{@{}c@{}}Time\\ (IGE)\end{tabular}} & \textbf{\begin{tabular}[c]{@{}c@{}}Time\\ (GL2vec)\end{tabular}} \\ \midrule
B1  & 3.230  & 2.280  & 3.378  & 3.461 \\
B2  & 5.812  & 2.668  & 7.317  & 69.491     \\
B3  & 9.101  & 3.866  & 42.268 & 636.969   \\
B4  & 5.365  & 6.518 & 13.048 & 10.688 \\
B5  & 7.143  & 10.569& 16.902 & 13.385    \\
B6  & 20.399 & 3.351  & 50.065 & 495.777  \\
B7  & 8.253  & 6.792  & 14.966 & 17.311   \\
B8  & 22.100 & 7.940 & 67.482 & 473.768    \\
M1  & 5.640  & 2.466 & 4.840  & 4.506 \\
M2  & 6.868  & 3.359 & 10.619  & 191.478  \\
M3  & 11.254 & 6.543 & 70.885  & 807.907  \\
M4  & 6.608  & 11.068  & 18.957  & 9.934 \\
M5  & 25.229 & 7.398 & 31.545  & 23.254   \\
M6  & 6.401  & 10.825  & 17.899  & 10.081 \\
M7  & 9.215  & 8.386 & 25.682  & 10.549  \\
M8  & 46.286 & 23.503 & 101.683  & 444.753  
\\ \midrule
\textbf{Avg.} & 12.432 &7.346 &31.096 &201.457   \\ \bottomrule
\end{tabular}
\end{adjustbox} 
\caption{Time complexity analysis.}
\label{Time_complexity}
\end{table}

\subsection{Graphs visualization with DHC-E features}
\begin{figure}[]
\centering
\includegraphics[scale=0.7]{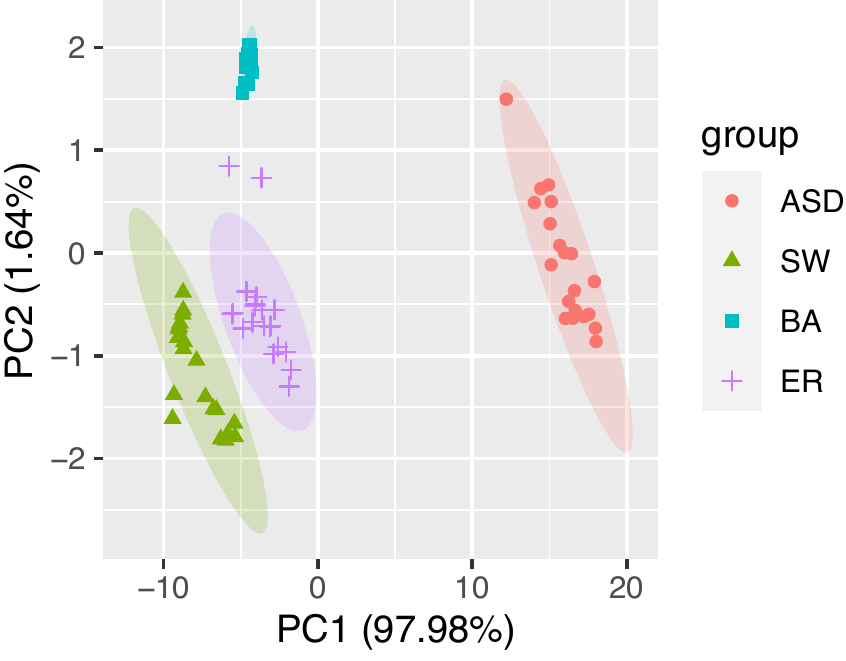}
\caption{\textbf{Lower-dimension morphospace representation of graph with DHC-E features}, we use the PCA to reduce the DHC-E whole graph embedding features into a two-dimension space and find that the DHC-E features can cluster together graphs in the same category. ASD, autism spectrum disorders' brain network; SW, small-world network; BA, scale-free network; ER, random network.}
\label{visGraph}
\end{figure}

We conducted an exploratory analysis in graph visualization applications to verify whether the DHC-E embedding method can adequately express different graphs in a two-dimensional space. First, we generated 20 small-world networks, 20 BA networks, and 20 random networks. Then, we combined these 60 networks with 20 ASD brain networks into a new dataset. To that end, we performed the DHC-E analysis on these 80 graphs and obtained the whole graph embedding for each graph. Subsequently, we performed principal component analysis (PCA) on the obtained embedded features to further reduce the embedding features to two dimensions. We observed that the first two PCs explain about 99.6\% of the total variability for the whole graph embedding. We found that it can cluster together graphs in the same category, which show the potential applications of DHC-E in lower-dimensional morphospace representation of graphs, as illustrated by Figure.~\ref{visGraph} in details.

\section{Conclusion and further consideration}
\label{Sec5}
In this study, we only consider the whole graph embedding rather than node or edge embeddings. We propose a simple, hyperparameter-free, and explainable graph embedding method and evaluate its performance on classification tasks using different datasets. We found that DHC-E is comparable but more straightforward than most sophisticated methods of graph embedding. By using DHC-E, we avoid problems such as tedious hyperparameters tuning and poor interpretability. As it stands, the DHC-E method shows good classification performance, stability and promising potential in lower-dimensional graph visualization. Additionally, DHC-E can be integrated with the current model to verify whether it can improve the classification performance and we built a brain network benchmark for further researches. Among researchers in interdisciplinary fields, a model that works out-of-the-box, has fewer parameters, and has moderately high performance may prove useful, and DHC-E may provide some guidance for future research into simple and effective whole graph embedding models.

\section{Data and code available}
\textbf{Supporting Information}, data and codes are available at \url{https://github.com/HW-HaoWang/DHC-E}
\section{Acknowledgements}
This work is supported by the National Natural Science Foundation of China (Nos. 61673150, 11622538), the Science Strength Promotion Program of the University of Electronic Science and Technology of China (No. Y030190261010020), and the China Scholarship Council (No. 201906070121).

\newpage
\bibliographystyle{named}
\bibliography{ijcai22}

\end{document}